\pdfoutput=1

\documentclass[11pt]{article}

\usepackage[]{acl2023}

\usepackage{times}
\usepackage{latexsym}

\usepackage[hang,flushmargin]{footmisc} %

\usepackage[T1]{fontenc}
\usepackage[utf8]{inputenc}

\usepackage{microtype}

\usepackage{inconsolata}

\title{AlbNews: A Corpus of Headlines for Topic Modeling in Albanian}

\author{Erion \c{C}ano \\
	Digital Philology \\
	Data Mining and Machine Learning \\
	University of Vienna, Austria \\
	\texttt{erion.cano@univie.ac.at} \\
	\And
	Dario Lamaj \\
	Cognitive Science \\
	Department of Applied Informatics \\
	Comenius University Bratislava \\
	\texttt{lamaj1@uniba.sk} \\
}

\begin{document}

\maketitle

\begin{abstract}
The scarcity of available text corpora for low-resource languages like Albanian is a serious hurdle for research in natural language processing tasks. This paper introduces AlbNews, a collection of 600 topically labeled news headlines and 2600 unlabeled ones in Albanian. The data can be freely used for conducting topic modeling research. We report the initial classification scores of some traditional machine learning classifiers trained with the AlbNews samples. These results show that basic models outrun the ensemble learning ones and can serve as a baseline for future experiments.   
\end{abstract}

\section{Introduction}

The AI (Artificial Intelligence) and LLM (Large Language Model) developments of today are creating a revolution in the way people solve many language-related tasks. Two pillars that paved the path to these developments were the Transformer neural network architecture \cite{NIPS2017_3f5ee243} and pretraining LLMs like BERT \cite{devlin-etal-2019-bert}. The internal mechanisms that drive the behavior and performance of LLMs are not fully known, but they resemble the human cognition mechanisms \cite{thoma-etal-2023-cogmemlm}. 

Other important factors have boosted AI applications been the increase of computing capacities \citep{DBLP:journals/corr/abs-2012-09303}, the development of high-level software frameworks \citep{10.5555/3454287.3455008}, and the creation of large text corpora. It is possible to use computing and software resources for solving tasks like Text Summarization, Sentiment Analysis, Topic Regnition etc., in any natural language. However, the language of the required text corpora must usually match the natural language under study. Most of the research corpora available today are in English. For underrepresented or low-resource languages like Albanian, few such corpora have been created and they are usually small in size. As pointed out by various studies \citep{blaschke-etal-2023-survey}, this is a severe limiation which hinders the performance and the variety of tasks that can be solved for low-resource languages. 

This paper presents AlbNews, a corpus of 600 labeled news headlines and 2600 unlabeled ones in Albanian.\footnote{Download from: \url{http://hdl.handle.net/11234/1-5411}} The headlines were collected from online news portals and the labeled part is annotated as \emph{pol} for ``politics'', \emph{cul} for ``culture'', \emph{eco} for ``economy'', and \emph{spo} for ``sport''. The unlabeled part of the corpus consists of headline texts only. The main purpose for creating and releasing this corpus is to foster research in topic modeling and text classification of Albanian texts. 
We performed some preliminary experiments on the labeled part of the corpus, trying a few traditional machine learning models.\footnote{Code at: \url{https://github.com/erionc/AlbNews}} The results we present can be used as comparison baselines for future experiments.

\section{Related Work} %

Topic Modeling is a common NLP research direction which develops AI techniques that can use unlabeled text samples for conducting topic analysis on text corpora. Text or topic classification, on the other hand, is slightly different and utilizes labeled text samples for training machine learning algorithms which are then used for predicting the topic of other text units or documents. In the context of this paper, we address both tasks, since AlbNews data can be used for both of them. 

Some of the erliest works on Topic Modeling such as Latent Semantic Indexing \citep{10.1145/275487.275505}, Probabilistic Latent Semantic Indexing \citep{10.1145/312624.312649}, and Latent Dirichlet Allocation \cite{10.5555/944919.944937} came out in late 90s and early 00s. They perform semantic similarity of documents based on word usage statistics. The most recent developments such as Top2Vec \cite{DBLP:journals/corr/abs-2008-09470} and BERTopic \cite{grootendorst2022bertopic} are based on LLMs and document embeddings. 

On the other hand, text classification has been traditionally conceived as a multiclass classification problem, given that the topic of a text unit can be one from a few predefined categories. It has been solved using traditional machine learning classifiers such as Support Vector Machine \cite{1174243}. Nevertheless, there are also recent studies that solve the text classification tasks using BERT or other LLMs \cite{DBLP:journals/corr/abs-1905-05583}. 

As for the resources, one of the most popular corpora is the 20 Newsgroups collection.\footnote{\url{http://qwone.com/~jason/20Newsgroups/}} It contains English texts of 20 different predefined topic categories such as \emph{politics}, \emph{religion}, \emph{medicine}, \emph{baseball}, etc. Many resources in English are based on copyleft texts of scientific \cite{meng-etal-2017-deep,topic-segmentation-2022,NIKOLOV18.2,cano-bojar-2019-efficiency}. They are large in size, but the samples are usually not annotated.  

In the context of low-resource languages, and more specifically Albanian, we are not aware of any corpus specifically created for Topic Modeling. There have been some similar attempts to create resources in Albanian for related NLP tasks. AlbMoRe is a recent corpus that contains 800 movie reviews in Albanian and the respective \emph{positive} or \emph{negative} sentiment annotations \citep{DBLP:journals/corr/abs-2306-08526}. It can be used to conduct research on sentiment analysis or opinion mining. Another similar resource is the collection of 10132 social media comments which were manually annotated and presented by \citet{KADRIU2022108436}. 

There have also been resources to solve other NLP tasks such as Named Entity Recognition. One of them is AlbNER, a collection of 900 sentences harvested from Albanian Wikipedia \cite{cano2023albner}. Each of its tokens has been manually annotated with the respective named entity tag. Finally, there are also resources like \emph{Shaj} corpus which is designed for hate speech detection \cite{DBLP:journals/corr/abs-2107-13592}. It contains user comments from various social media, annotated using the OffensEval schema for hate speech detection.\footnote{\url{https://sites.google.com/site/offensevalsharedtask}}

\section{AlbNews Corpus} %

\begin{table}
\centering
\begin{tabular}{l c c}
\hline
& \textbf{Characters} & \textbf{Tokens}\\
\hline
Minimum & 38 & 5 \\
Maximum & 187 & 33 \\
Average & 88.13 & 13.98 \\\hline
\end{tabular}
\caption{Headline length statistics.}
\label{tab:lenth}
\end{table}

Driven by the increases in advertising revenew, news websites and news pages in social media are becoming the biggest sources of everyday information nowadays.\footnote{\url{https://www.pewresearch.org/journalism/fact-sheet/digital-news/}} They provide information of various topics such as \emph{politics}, \emph{economy}, \emph{sport}, \emph{fashion}, \emph{culture}, \emph{technology}, etc. The essence of the news comes from the headlines which are typically 1 to 3 sentences long.  

AlbNews was created by collecting such headlines of Albanian news articles. Each article was published during the period February 2022 - December 2023. Initially, about 6000 headlines were crawled. Some of them were droped, since they were very short or had badly-formatted content. Only headlines consisting of at least one full and properly formatted sentence were kept. At the end, a total of 3200 headlines was reached. The length statistics in characters and tokens for these final 3200 headlines are shown in Table~\ref{tab:lenth}.

Because of the limited manpower, only 600 from the 3200 headlines were randomly selected for annotation. The first and the second author worked separately and labeled each headline as \emph{pol} for ``politics'', \emph{cul} for ``culture'', \emph{eco} for ``economy'', and \emph{spo} for ``sport''. In most of the cases, the two labels from each annotator matched. The few cases of missmatches were resolved through discussion. Table~\ref{tab:samples} presents four labeled samples, one per each category.

\begin{table}
\centering
\begin{tabular}{p{5.5cm} | c }
\hline
\textbf{Headline} & \textbf{Topic}\\
\hline
Zgjedhjet vendore: rreth 230 vëzhgues të huaj në Shqipëri & pol \\\hline
Lahuta në UNESCO, komuniteti i bartësve propozon masat mbrojtëse & cul \\\hline
Ulet me shtatë lekë nafta! Çmim më i lirë dhe për benzinën e gazin & eco \\\hline
Spanja godet Italinë në minutën e fundit dhe shkon në finale & spo \\\hline
\end{tabular}
\caption{Illustration of four data samples.}
\label{tab:samples}
\end{table}

\section{Preliminary Experimental Results} \label{sec:results}

In this section, we present some preliminary results obtained using the AlbNews corpus. We trained a few traditional machine learning algorithms on the labeled part of the corpus and observed their performance on the topic classification task. 

\subsection{Preprocessing and Vectorization} \label{ssec:preproc}

We performed some preprocessing steps on each headline text. They were first tokenized and the punctuation or special symbols were removed. The white-space symbols like `\verb|\n|' or `\verb|\t|' were also lost. The texts were also lowercased to decrease the vocabulary (unique words). At the end, TF-IDF was applied to vectorize the text words \cite{ZHANG20112758}. These preprocessing operations are not related to the semantics of the words and do not usually have any influence in topic classification performance.   

\subsection{Classification Algorithms} \label{ssec:algorithms}

One of the most successfull algorithms created in the 90s is SVM (Support Vector Machine). It has revealed itself to be fruitfull in both classification and regression tasks \cite{cortes1995support}. SVM is based on the notion of hard and soft class separation margins and was subsequently improved by adding the \emph{kernel} concept which makes it possible to separate data that are not linearly separable by means of feature space transformations \cite{Kocsor:2004:AKF:1008633.1008643}.

Another simple but effective classifier is Logistic Legression which offers good performance on a wide range of applications. Logistic Regression utilizes the logistic function for determining the probability of samples belonging to certain classes. Moreover, it runs quite fast.
Decision trees utilize hierarchical tree structures to analyze the data features (appearing in tree branches) and make decisions (tree nodes) accordingly. They have been around since many years and have shown strong classification performance when applied on data of different types \cite{quinlan:induction}. 

Other successful classification models are those which combine several basic algorithms to create stronger ones. This concept is knowns as ensemble learning \cite{Brown2010}. One family of ansemble learners is that which is based on the boosting concept \cite{Schapire2003}. They try to find prediction ``rules of thumb'' using basic models, and improving the process by repeatedly feeding different training samples on each model instance. After a certain number of iterations, the boosting algorithm combines the learned rules into a single prediction rule that usually is more accurate. Two implementations of this method are Gradient Boosting \cite{10.1214/aos/1013203451} and XGBoost \cite{10.1145/2939672.2939785}.  

A different approach to ansemble learning is called bagging and tries to generate multiple versions of a model \cite{10.1023/A:1018054314350}. The predictions of the multiple predictors are aggregated to provide the final predictions. One implementation of this idea is the Random Forest method \cite{Ho:1995:RDF:844379.844681}. It aggregates predictions obtained from several decision trees.  

\begin{table}
\centering
\begin{tabular}{l c}
\hline
\textbf{Model} & \textbf{Accuracy} \\
\hline
Logistic Regression & 0.85 \\
Support Vector Machine & 0.841 \\
Decision Trees & 0.5 \\
Gradient Boosting & 0.683 \\
XGBoost & 0.541 \\
Random Forest & 0.641 \\\hline
\end{tabular}
\caption{Topic classification results.}
\label{tab:results}
\end{table}

\subsection{Discussion} \label{ssec:discussion}

We trained the classifiers mentioned above with their default parameters on 480 labeled samples and tested on the remaining 120 samples. The accuracy results that were reached are shown in Table~\ref{tab:results}. As we can see, Logistic Regression gives an accuracy score of 0.85 which is the highest. SVM follows, reaching up to 0.84. Decision trees are significantly weaker, reaching only 0.5 (note that random guessing on four categories yelds an accuracy score of 0.25). Random forest performs slightly better, with an accuracy score of 0.64. Even Gradient Boosting and XGBoost, the two boosting implementations do not perform well. They reach up to 0.68 and 0.54 only. 

The results indicate that the simpler methods outrun the more advanced ensemble learning ones. One reason for this could be overfitting which happens often when the data are small. Another possibility would be to grid-search and optimize some of the parameters of the classifiers. This could somehow boost their scores, but is beyond the scope of this work and could be a future work extension. Another possible future work could be experimenting with LLMs such as BERT or RoBERTa \cite{DBLP-roberta}. However, there are still no such LLMs pretrained on Albanian texts. Knowledge transfer between English and Albanian leads to poor results \cite{cano2023albner}, highligting the need for developing LLMs pretrained with Albanian texts.

\section{Conclusions} \label{sec:conclusions}

Experimenting on NLP tasks such as Topic Modeling demands the creation of unsupervised, semi-supervised or supervised corpora which in the cases of low-resource languages like Albanian are unavailable, scarce or small. This work introduces AlbNews, a collection of news headlines in Albanian. It conists of 600 labeled headlines and 2600 unlabeled headlines and aims to foster research on Topic Modeling of Albanian texts. A set of preliminary experiments with tradicional machine learning methods indicates that the simple ones perform better than those based on ensemble learning.   

\bibliography{custom}
\bibliographystyle{acl_natbib}

\end{document}